\documentclass[a4paper,12pt,onecolumn]{article}

\usepackage{FAS}
\usepackage[utf8]{inputenc}
\usepackage{graphicx}

\usepackage{xcolor}
\usepackage{amsmath,amsfonts}
\usepackage{algorithmic}
\usepackage{algorithm}
\usepackage{array}
\usepackage{textcomp}
\usepackage{stfloats}
\usepackage{url}
\usepackage{verbatim}
\usepackage{cite}

\usepackage{hyperref}
\usepackage[acronyms,toc,nonumberlist,nogroupskip]{glossaries}
\glsdisablehyper
\glsnoexpandfields
\usepackage{cleveref}

\newacronym{ad}{AD}{Automated driving}
\newacronym{ads}{ADS}{\gls{ad} system}
\newacronym{ade}{ADE}{average distance error}
\newacronym{bev}{BEV}{birds eye view}
\newacronym{ecdf}{ECDF}{empirical cumulative distribution function}
\newacronym{nds}{NDS}{nuScenes detection score}
\newacronym{ooi}{OOI}{object of interest}
\newacronym{pkl}{PKL}{planning Kullback-Leibler divergence}
\newacronym{rss}{RSS}{responsibility sensitive safety}
\newacronym{ttc}{TTC}{time to collision}
\newacronym{sacred}{SACRED}{Stuctured Analysis for Conservative Relevance Estimation in Driving context}
\newacronym{sure}{SURE}{Safe Urban Relevance Extension}

\usepackage{subcaption}

\begin{document}

\title{SURE-Val: \textbf{S}afe \textbf{U}rban \textbf{R}elevance \textbf{E}xtension and \textbf{Val}idation}
%
%
\author{Kai Storms\footnote{contributed equally} \footnote{Institute of Automotive Engineering (FZD) at Technical University of Darmstadt, 64287 Darmstadt (e-mail: firstname.lastname@tu-darmstadt.de)}
%
\ ,Ken Mori\footnotemark[1] \footnotemark[2]
%
\ und Steven Peters\footnotemark[2]}
%
%
\date{}

\maketitle \thispagestyle{empty}


\begin{abstract}
To evaluate perception components of an automated driving system, it is necessary to define the relevant objects. 
While the urban domain is popular among perception datasets, relevance is insufficiently specified for this domain.
Therefore, this work adopts an existing method to define relevance in the highway domain and expands it to the urban domain. 
While different conceptualizations and definitions of relevance are present in literature, there is a lack of methods to validate these definitions.
Therefore, this work presents a novel relevance validation method leveraging a motion prediction component.
The validation leverages the idea that removing irrelevant objects should not influence a prediction component which reflects human driving behavior.
The influence on the prediction is quantified by considering the statistical distribution of prediction performance across a large-scale dataset. 
The validation procedure is verified using criteria specifically designed to exclude relevant objects. 
The validation method is successfully applied to the relevance criteria from this work, thus supporting their validity. 

\end{abstract}


\section{Introduction}

\gls{ad} is currently viewed as a key emerging technology. 
Among the benefits which are attributed to \gls{ad} are increased comfort, availability of mobility and foremost an increase in safety for the traffic environment~\cite{Canis.23.05.2017}. 
However, safety assurance of \gls{ad} remains a challenge.
Whilst first SAE level 3 \gls{ads} are already available~\cite{drivepilot} for parts of the highway domain, a current focus in research is the urban domain~\cite{vvm}.

One method supporting safety assurance by offering potential benefits regarding the testing effort is modular decomposition~\cite{Amersbach.}.
Under this method, all modules of a classic Sense-Plan-Act architecture and its derivatives, such as the perception module, need to be evaluated individually~\cite{philipp2020functional}. 
For this individual evaluation, it is imperative to have knowledge about what is relevant to the subject module, both for completeness and efficiency.

Therefore, in this paper we will consider relevance for the perception module and how it can be evaluated.
Current approaches face a variety of problems.
Most methods lack generality because they are specific to either one module or a limited set of scenarios. 
These methods cannot be applied to other modules or scenarios without requiring major redesign efforts.

Furthermore, a lack of consistency between relevance results must be noted among different methods. 
One reason is the lack of validation in previous methods. 
This situation is exacerbated by the fact that the current state of the art does not provide any methodology to validate relevance criteria.

This paper will expand on the previous work of  Mori \& Storms\cite{MoriStorms2023arxiv}, henceforth denoted as \gls{sacred}.
Firstly, the \gls{sure} is introduced.
As main contribution, we present a novel methodology that enables the evaluation of validity for a given relevance selection method.
This methodology is applied and verified using the expanded method for the urban domain.

\section{Related Work}

This section covers related works with respect to concepts of relevance as well as their respective validation.

\subsection{Relevance} 

Previous conceptualizations of relevance can be broadly categorized in heuristic approaches, formal approaches and concepts based on a downstream task. 

Heuristic approaches typically implicitly define relevance by excluding certain objects based on simple criteria.
One application is the datasets used to develop and test perception functions.
Here, thresholds on distance~\cite{Sun.2019} or number of points from lidar and radar~\cite{Caesar.2020} are applied during annotation. 
Heuristics are also applicable when defining perception metrics~\cite{Hoss.2022}.
Criteria such as distance~\cite{Caesar.2020}, height in camera image plane and occlusion~\cite{Geiger.2012} are used by dataset metrics.
Other metric proposals in literature follow similar approaches such as leveraging the distance to the ego vehicle~\cite{Lyssenko.2021} or the ego trajectory~\cite{Berk.2020}. 
Similar ideas are found in neural path planners where relevance is implicitly defined by the network input.
Commonly, inputs are restricted to a certain geometric region~\cite{Bansal.2018, Sadat.2020, Philion.2020, Hallerbach.2018} or a limited number of objects~\cite{Xu.2018, Cho.2019, Ettinger.2021, Houston.2020, Vazquez.2022}. 

Formal approaches provide an explicit consideration of relevance based on requirements. 
Typically, relevance is related to safe behavior of the vehicles by considering reachability or formal planners~\cite{Hoss.2022}. 
Reachability leverages kinematic constraints to define potential collision objects as relevant~\cite{Althoff.2010, Topan.2022}. 
Other work leverages formal planners either from preexisting work~\cite{Volk.2020} or by directly specifying context-dependent behavioral requirements~\cite{Schonemann.2019, Philipp., MoriStorms2023arxiv} 

In order to avoid manual specification of relevance, the \gls{pkl}~\cite{Philion.2020b} and following work~\cite{Henze.2021, Philipp.} propose to leverage neural planners. 
Here, relevance is conceptualized and quantified as magnitude of the effect of an object on a downstream planner implementation~\cite{Philion.2020b}.
However, the validity of this approach is limited to the specific implementation of the planning algorithm~\cite{Philipp.2021}.

Overall, various approaches for the definition of relevance are proposed and reach different conclusions.
Notably, there is currently no reconciliation of formal and downstream implementation based approaches.

\subsection{Relevance Validation}

While different approaches have been suggested to conceptualize object relevance, the validation of these concepts is currently underexplored. 
Perception datasets such as~\cite{Geiger.2012, Sun.2019, Wilson.2021} do not consider the validity of the perception metrics with respect to safety.
Other datasets discuss metrics with respect to the weighting of different attributes~\cite{Mao.2021} or validate the ability of the metric to produce a ranking between different types of detectors~\cite{Caesar.2020}

Some analytic approaches rely exclusively on plausibilization by visualizing selected scenarios without further validation~\cite{Topan.2022, Volk.2020, Philipp.}. 
In other cases, no further attempts to verify or validate the results are made~\cite{Deng.2021}. 

The usage of planners in a closed loop simulation has been proposed~\cite{Topan.2022} and also executed for safety aware prediction metrics~\cite{Jha.2022}.
However, incorporating a planner into the pipeline incurs the potential of errors in the planner which questions the validity~\cite{Mao.2023}.
The results of the \gls{pkl} metric are verified and validated in two different ways. 
A first verification step argues plausibility by showing that the metric is sensitive to distance and velocity as intuitively salient features. 
Furthermore, a human evaluation is conducted by Amazon Mechanical Turk workers, showing an 80\% preference for the \gls{pkl} metric over the \gls{nds}.

Overall, current validation approaches for validating relevance criteria are lacking. 
Furthermore, the suggested approaches generally focus on relative ranking and therefore do not provide acceptance criteria. 

\section{Methodology} 

We suggest the following approach for defining and validating relevance in this work.
First, a formal analytic method of relevance is selected to facilitate interpretability.
Next, a neural trajectory prediction model is utilized to validate the results of the formal model. 
It will later be shown that neural networks are suitable for validation if large-scale datasets are leveraged to consider uncertainties in the prediction outcome. 

As formal relevance method, the prior \gls{sacred} method\cite{MoriStorms2023arxiv} is selected. 
This method provides a conservative estimation aiming to provide a complete set of relevant objects. 
The property of completeness will also be focus of the later validation procedure. 
In addition, the relevance method offers the benefit of few requirements regarding environment knowledge.

\section{Application of Relevance Selection Method}

This section follows the \gls{sacred} relevance method defined in \cite{MoriStorms2023arxiv}.
First, a minimum specification is provided.
Next, interactions are decomposed into multiple scenarios for each of which relevance criteria are formulated. 

\subsection{Specification}

In order to apply the relevance methodology, it is first necessary to specify the system and the use case. 

The system specification follows \gls{sacred}~\cite{MoriStorms2023arxiv}.
A Sense-Plan-Act architecture is assumed with an object list as interface between perception and planning. 
The output trajectories of the planner must generally adhere to physical and legal requirements. 
This work focuses on the task of collision safety. 
In order to fulfill these requirements, the system provides guarantees regarding its capabilities.
After a specified reaction time, the system is required to act in accordance with the requirements. 
In order to do so, the system is capable of providing a minimum guaranteed braking deceleration as well as a minimum guaranteed acceleration.

Currently, testing of perception is mainly performed on datasets such as \cite{Caesar.2020} which mainly consider urban environments. 
Therefore, the use case is selected to be the application in the urban domain. 
This use case serves to expand the ideas from the highway domain to more complex interactions.

\subsection{Decomposition} 

The next step in determining relevance is the decomposition of the use case into functional scenarios. 
As in \gls{sacred}~\cite{MoriStorms2023arxiv}, the distinction is made using polar coordinates for the relative velocities, distinguishing radial and tangential scenarios (first letter of abbreviation). 
Depending upon whether the ego is moving towards or away from the \gls{ooi}(second letter of abbreviation) and whether the \gls{ooi} is moving towards or away from the ego (third letter of abbreviation), the four radial scenarios R.TA, R.AT, R.TT and R.AA are distinguished. 
The tangential scenarios are distinguished depending on whether the \gls{ooi} is moving towards or away from the ego vehicle.
This is described by the two scenarios T.XT and T.XA. 

The results for the R.TA, R.AT, R.AA and T.XA scenario are identical to \gls{sacred}~\cite{MoriStorms2023arxiv} and are thus not repeated here.
However, the R.TT and the T.XT scenario require additional considerations to consider the urban environment.
In order to operate in an urban environment a vehicle needs to be able to pass static obstacles by entering the opposite lane and to traverse intersections.
Overtaking dynamic objects is excluded to limit complexity, since it is not required to fulfill a driving task.
These are considered by modifying the R.TT and the T.XT scenario respectively.
Each of the scenarios requiring modification to the relevance estimation is discussed in the following subsections.

\subsection{R.TT'}

First, the R.TT scenario is applied as in \gls{sacred}~\cite{MoriStorms2023arxiv} without modification.
For all objects not considered relevant by the R.TT scenario, a modified R.TT scenario is used.
This scenario, shown in Fig.~\ref{fig:passing}, hereafter called R.TT', takes into account the case of passing a static object. 

Variables follow the notation from \gls{sacred}~\cite{MoriStorms2023arxiv} with three optional indices.
The first index identifies the object, with 1, 2 and 3 denoting the ego vehicle, opposing vehicle and static object respectively.
The second index gives the frame of reference, either r for radial or t for tangential.
The third index denotes the state.


We extend the \gls{sacred} method~\cite{MoriStorms2023arxiv} which only considers pairwise interactions.
First, the static object is considered as \gls{ooi}.
Since the \gls{ooi} is static, the R.TT and R.TA are equivalent.
In order to distinguish the R.TT' scenario, the static object has to be relevant according to either scenario.
Arbitrarily selecting R.TA requires the distance to the static object $d_\mathrm{1,r,0}$ to be less than the sum of reaction and breaking distance:



\begin{figure}
\centering
\includegraphics[width=0.85\linewidth]{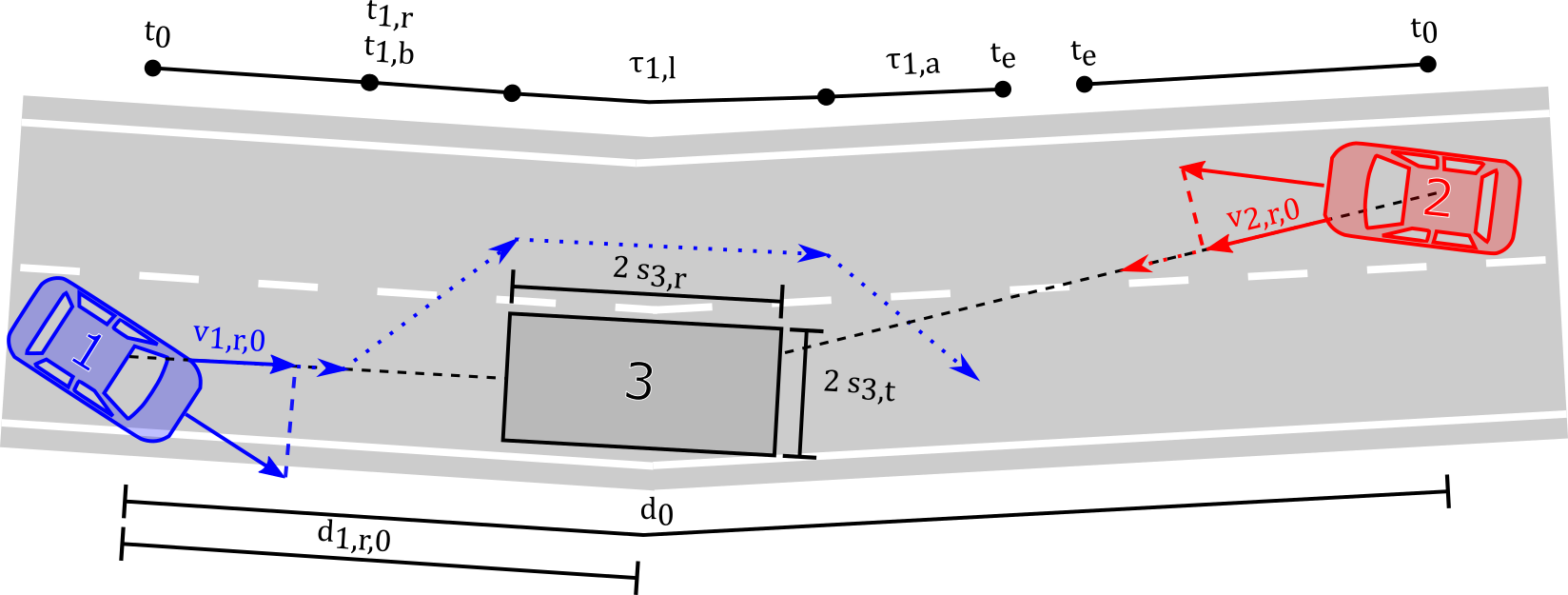} 
\caption{Sequence and variables for the R.TT' scenario.}
\label{fig:passing}
\end{figure}

\begin{equation}
    0<d_\mathrm{1,r,min} = d_\mathrm{1,r,0}  - s_\mathrm{1} - s_\mathrm{3,r} \\
    -   v_\mathrm{1,r,0} \ t_\mathrm{1,r} + \frac{1}{2} a_\mathrm{max}  t_\mathrm{1,r}^2 \\
    - \frac{(v_\mathrm{1,r,0} + t_\mathrm{1,r} a_\mathrm{max})^2}{2 a_\mathrm{1,r,b}} 
\end{equation}

The R.TT' scenario is applied if all following conditions for the initial positions $r_\mathrm{i,0}$ and velocities $v_\mathrm{i,0}$ are fulfilled:

    \begin{itemize}
    \item Ego moving towards static object: $(\vec{r}_\mathrm{3,0} - \vec{r}_\mathrm{1,0}) \cdot \vec{v}_\mathrm{1,0} > 0$
    \item \gls{ooi} moving towards static object: $(\vec{r}_\mathrm{3,0} - \vec{r}_\mathrm{2,0}) \cdot \vec{v}_\mathrm{2,0} > 0$
    \item \gls{ooi} located behind static object: $(\vec{r}_\mathrm{3,0} - \vec{r}_\mathrm{1,0}) \cdot (\vec{r}_\mathrm{3,0} - \vec{r}_\mathrm{2,0}) < 0$
\end{itemize}

The scenario is described by the following considerations, according to worst case assumptions.
Within the reaction time, the ego vehicle brakes.
This results in a distance covered and velocity as defined in \eqref{eq:d1b} and \eqref{eq:v1b}.
The end time of the braking maneuver $t_\mathrm{1,b}$ in \eqref{eq:t1b} is equal to the reaction time or the time to standstill if less.

\begin{equation}\label{eq:d1b}
    d_\mathrm{1,b} =  v_\mathrm{1,r,0} t_\mathrm{1,b} - \frac{1}{2} a_\mathrm{max}  t_\mathrm{1,b}^2 \
\end{equation}

\begin{equation}\label{eq:v1b}
    v_\mathrm{1,r,b} = v_\mathrm{1,r,0} - a_\mathrm{max} t_\mathrm{1,b}
\end{equation}

\begin{equation}\label{eq:t1b}
    t_\mathrm{1,b} = min \biggl\{ t_\mathrm{1,r}\ , \ \frac{v_\mathrm{1,r,0}}{a_\mathrm{max}} \biggr \}
\end{equation}

After the reaction time, the passing maneuver is initiated with a lateral movement to the opposite lane, followed by accelerating with the guaranteed acceleration.
After passing the static object, the ego vehicle performs a lateral movement back to its initial lane, thus concluding its maneuver.
The durations of the lateral lane change $\tau_\mathrm{1,l}$ and acceleration $\tau_\mathrm{1,a}$ are defined in \eqref{eq:tau1l} and \eqref{eq:tau1a}.
The final velocity after accelerating $v_\mathrm{1,a}$ is given by \eqref{eq:v1a}.

\begin{equation}\label{eq:tau1l}
    \tau_\mathrm{1,l} = 2 \sqrt{\frac{s_\mathrm{3,t}+s_\mathrm{1}}{a_\mathrm{1,g}}}
\end{equation}



\begin{equation}\label{eq:v1a}
    v_\mathrm{1,r,a} = v_\mathrm{1,r,b} + a_\mathrm{1,g} \tau_\mathrm{1,a}
\end{equation}

\begin{equation}\label{eq:tau1a}
    \tau_\mathrm{1,a} = \frac{-v_\mathrm{1,r,b} + \sqrt{2 a_\mathrm{1,g} (2 s_\mathrm{1} + 2s_\mathrm{3,r}) + v_\mathrm{1,r,b}^2}}{a_\mathrm{1,g}}
\end{equation}

The final distance the ego vehicle traverses during the scenario is defined in \eqref{eq:d1e} as sum of the individual maneuvers.

\begin{equation}\label{eq:d1e}
    d_\mathrm{1,e} = 
    d_\mathrm{1,b} + v_\mathrm{1,r,b}\tau_\mathrm{1,l} + d_\mathrm{1,a} + v_\mathrm{1,r,a}\tau_\mathrm{1,l} 
\end{equation}

Throughout the scenario, the opposing vehicle performs the maximum acceleration towards the ego, yielding \eqref{eq:d2e} and \eqref{eq:v2e} with \eqref{eq:te}.

\begin{equation}\label{eq:d2e}
    d_\mathrm{2,e} = v_\mathrm{2,r,0} t_\mathrm{e} + \frac{1}{2} a_\mathrm{max} t_\mathrm{e}^2
\end{equation}

\begin{equation}\label{eq:v2e}
    v_\mathrm{2,r,e} = v_\mathrm{2,r,0} + a_\mathrm{max} t_\mathrm{e}
\end{equation}

\begin{equation}\label{eq:te}
    t_\mathrm{e} = t_\mathrm{1,r} + 2\tau_\mathrm{1,l} + \tau_\mathrm{1,a}
\end{equation}

Given the resulting distances and velocities of the defined events, the relevance of the opposing vehicle can then be determined by using them as 
the initial values for R.TT from \gls{sacred}~\cite{MoriStorms2023arxiv} as defined in equation \eqref{eq:DistanceRTT}.

\begin{equation} \label{eq:DistanceRTT}
\begin{aligned}
    0 < d_\mathrm{min} = 
    & \ (d_\mathrm{0} - d_\mathrm{1,e} - d_\mathrm{2,e}) - s_\mathrm{1} - s_\mathrm{2} - v_\mathrm{1,r,a} \ t_\mathrm{1,r} \\
    & - \frac{1}{2} a_\mathrm{max}  t_\mathrm{1,r}^2 - \frac{(v_\mathrm{1,r,a} + t_\mathrm{1,r} a_\mathrm{max})^2}{2 a_\mathrm{1,r,b}} 
    - v_\mathrm{2,r,e} t_\mathrm{1,b} - \frac{1}{2} a_\mathrm{max} t_\mathrm{1,b}
\end{aligned}
\end{equation}

Exemplary scenarios yield distances in the hundreds of meters, well beyond typical dataset annotation ranges.
In addition, passing only needs to be considered given an ego intention.
Since this information is not available on datasets, this scenario is neglected for the practical implementation.

\subsection{T.XT'}

This scenario covers tangential movements such as merging and intersections. 
The merging covered in \gls{sacred}~\cite{MoriStorms2023arxiv} is extended to intersections encountered in the urban domain. 

\begin{figure}
\centering
\includegraphics[width=0.75\linewidth]{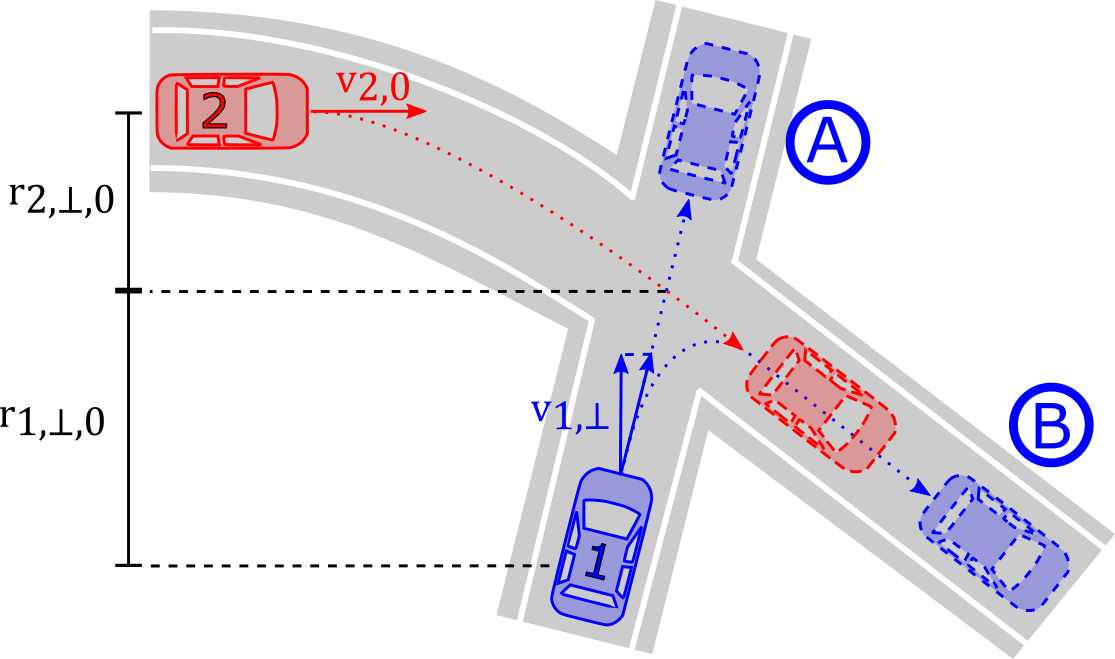} 
\caption{Worst-case intersection and variables in the T.XT' scenario.}
\label{fig:intersection}
\end{figure}

Within an intersection, there are two distinct maneuvers available to the ego vehicle  as shown in Fig.~\ref{fig:intersection}. 
It can either merely pass through (A) or turn onto the path of another vehicle travelling (B).
The requirement in either case is not to impede another vehicle.
The latter case represents the worst case since ego vehicle must not only leave the intersection, but additionally accelerate to an adequate speed. 

Thus, the worst case in an intersection is adequately described by the T.XT scenario from \gls{sacred}~\cite{MoriStorms2023arxiv}. 
The main difference is that the lateral distance to the travelling direction of the other vehicle can be arbitrarily large unlike for the case of the highway domain. 
Therefore, the ego vehicle has the opportunity to avoid entering the intersection by braking in time. 
The corresponding braking distance is: 

\begin{equation}
    s_\mathrm{1,\bot,b} = v_\mathrm{1,\bot,0} t_\mathrm{1,r} + \frac{v_\mathrm{1,\bot,0}^2}{2 a_\mathrm{1,\bot,b}}
\end{equation}

Additionally, we assume the location of the intersection to be unknown.
Therefore, the intersection may be closer to the ego than the heading of the other vehicle. 
In the worst case, the other vehicle would perform a maximum lateral acceleration towards the ego vehicle to reach the intersection. 
The corresponding lateral distance traveled with the corresponding braking time of the ego vehicle is:

\begin{equation}
    d_\mathrm{2,\bot} = \frac{1}{2} a_\mathrm{max} t_\mathrm{1,b}^2 
\end{equation}
\begin{equation}
    t_\mathrm{1,b} = t_\mathrm{1,r} + \frac{v_\mathrm{1,\bot,0} + t_\mathrm{1,r} a_\mathrm{max} }{a_\mathrm{1,b}} 
\end{equation}

Considering these two components, the maximum lateral distance for which the merging scenario from \gls{sacred}~\cite{MoriStorms2023arxiv} is considered is limited to:

\begin{equation}
    r_\mathrm{1,\bot,0} < 
    v_\mathrm{1,\bot,0} t_\mathrm{r} + \sqrt{\frac{v_\mathrm{1,\bot,0}^2}{2 a_\mathrm{1,\bot,b}}} +
    \frac{1}{2} a_\mathrm{max} \left[t_\mathrm{1,r} + \frac{v_\mathrm{1,\bot,0} + t_\mathrm{1,r} a_\mathrm{max} }{a_\mathrm{1,b}} \right]^2 
\end{equation}

If this equation is fulfilled, the merging scenario from \gls{sacred}~\cite{MoriStorms2023arxiv} is applied.
Otherwise, the respective object is considered irrelevant according to this scenario.

\section{Validation}

This section discusses the validation of the results presented thus far. 
First, some preliminary discussion relating to prior work is shown.
Next, the approach of this work is presented.
Finally, the approach is applied and verified. 

\subsection{Preliminaries}

As noted in previous sections, the validation of relevance criteria has received limited attention with no generally accepted methodology available. 
Previous relevance concepts either take an argumentative approach or include a concrete downstream planning task. 
In addition, the notion of a human baseline is present in one work~\cite{Philion.2020b}.
We believe that reconciliation of these complimentary approaches is required in order to argue validity.

One basis for the proposed validation approach is the human baseline. 
This idea has previously been applied for accident rates of human driving performance~\cite{PEGASUSProject.2019, Junietz.2019, Liu.2019} as well as for perception performance~\cite{Qi.2021}.
Works incorporating downstream planners implicitly also include this concept.
However, planning goals may include additional aspects other than imitating human behavior~\cite{Bansal.2018} such as infractions, mission goals~\cite{Dosovitskiy.2017} and comfort~\cite{Caesar.2021}.
Therefore, these goals are ambiguous and lack consistent evaluation  metrics~\cite{Caesar.2021}. 

\subsection{Proposed Method} 

We conceptualize relevance validation by considering the human behavior as baseline. 
Ideally, the influence of removing objects considered irrelevant might be studied in driving simulators to directly quantify the behavioral changes in human subjects. 
To avoid the substantial costs such an approach incurs, we propose to approximate human driving behavior with a motion prediction algorithm. 
This approach follows the approach proposed in~\cite{MoriStorms2023arxiv} which is adapted from~\cite{Philion.2020b}.
A motion prediction algorithm has several advantages over using a path planning algorithm. 
The objective of the motion prediction is unambiguous and can be evaluated in an open-loop setting with real-world data without relying on a simulation environment. 
This work uses the predictor as proxy for human behavior rather than as component to create a full \gls{ad} pipeline as in~\cite{Philion.2020b}.
Additionally, this work explicitly quantifies the prediction performance as opposed to prior work which explicitly or implicitly assumes the planner to be valid~\cite{Philion.2020b, Jha.2022}.

The validation procedure consists of running the prediction network with different inputs. 
Predictions are calculated multiple times for each input to account for potential uncertainties and non-deterministic elements of the prediction. 
The validation procedure only considers the \gls{ecdf}  of prediction errors across a large-scale dataset to avoid local performance issues and effects of non-deterministic predictions.
A relevance criterion is considered valid if the prediction error distribution remains unchanged between the original input and the filtered input.
Whether two distributions are identical is evaluated using the Cramer-von Mises test for two empirical distributions\cite{Anderson.1962}.
This test provides a confidence value, based on distribution similarity and sample size, for which an acceptance criterion is required.

To study the utility of the proposed validation method, an implementation on the nuScenes dataset~\cite{Caesar.2020} is performed.
All experimental results are reported for the standard val split. 
The prediction network is selected from the nuScenes motion prediction leaderboard~\cite{nuScenes.2020} among entries with an open implementation. 
We select the PGP algorithm~\cite{nachiket92.2022} as implemented by~\cite{Deo.2022} using the default settings. 
The implementation filters inputs by location in a square region from [-20~m, 80~m] in longitudinal and [-50~m,~50~m] in lateral direction. 
This filtering is maintained for all experimental conditions with the relevance filtering criteria only being additionally applied. 
The prediction errors are evaluated using the \gls{ade} metric for the top 10 trajectories.

\subsection{Results}

\begin{figure}
\begin{minipage}{.48\textwidth}
  \raggedright
  \includegraphics[width=0.95\textwidth]{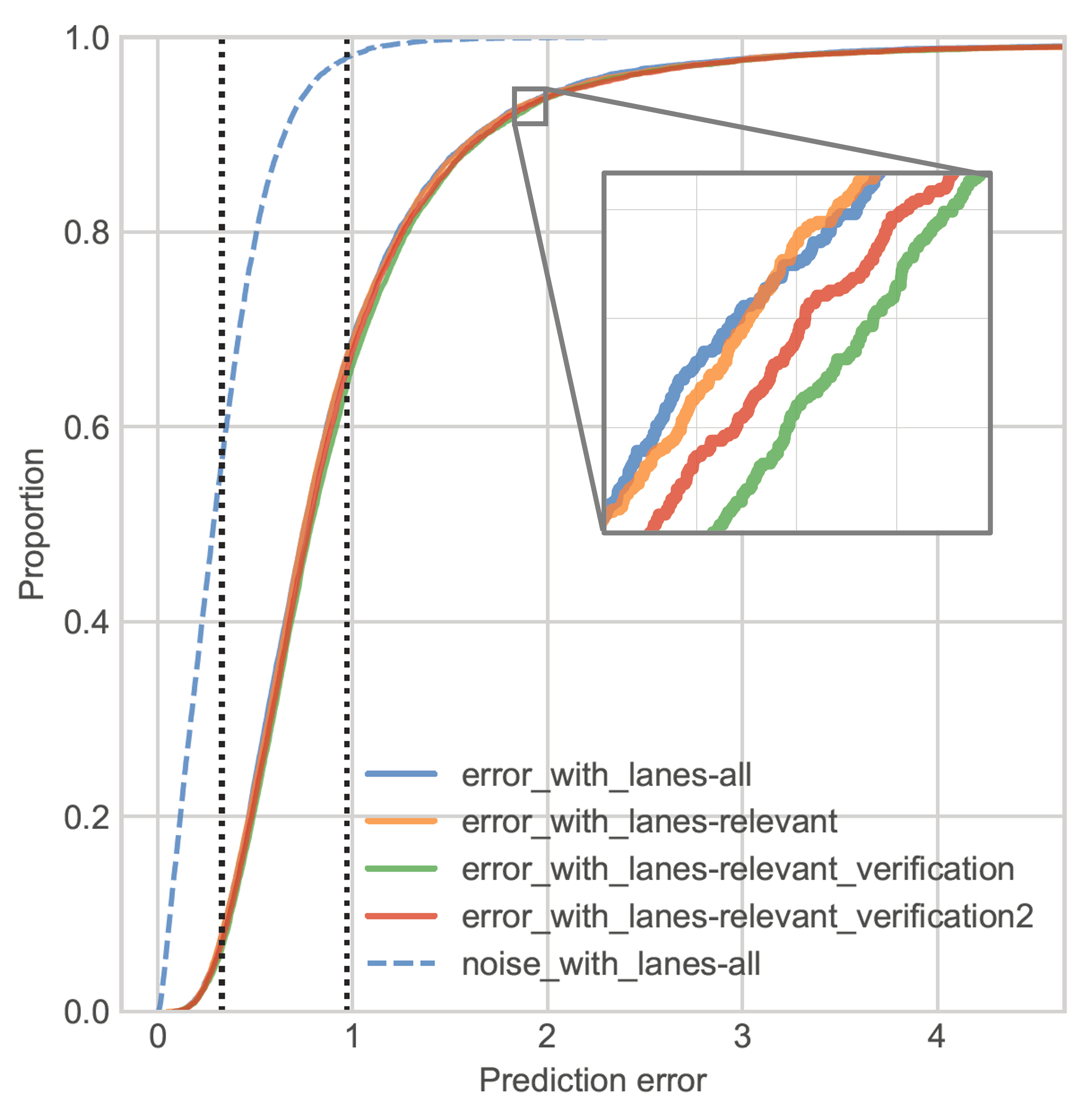}
  \captionof{figure}{\gls{ecdf} prediction error as \gls{ade} for top 10 trajectories for different inputs and noise for multiple runs with same inputs.}
  \label{fig:distribution_errors}
\end{minipage}%
\hspace{.04\textwidth}
\begin{minipage}{.48\textwidth}
  \raggedleft
  \includegraphics[width=\textwidth]{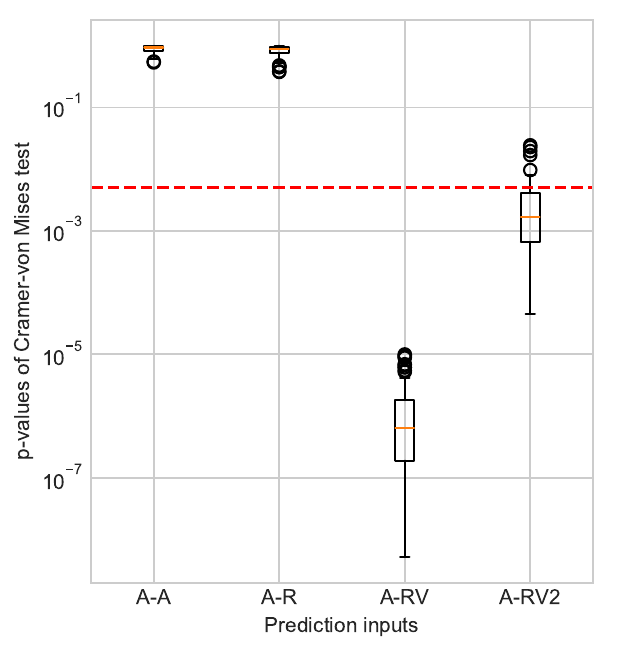}
  \captionof{figure}{Boxplot of p-values testing for equality of distribution for pairwise combinations. Exemplary threshold indicated by red dashed line.}
  \label{fig:test_conf}
\end{minipage}
\end{figure}

The prediction network is run ten times each with four different inputs. 
Firstly, multiple prediction runs with all inputs abbreviated as `A' are compared to each other to determine the prediction  noise.
The \gls{ecdf} of this noise is depicted in Fig.~\ref{fig:distribution_errors}, showing that the detector exhibits significant noise averaging at 0.33~m. 
The error distribution is depicted as \gls{ecdf} averaging at 0.96 m.
Noise also affects the error distributions.
Therefore, different runs using identical inputs with all objects (A-A) are compared using statistical tests for equality of distribution.
The results are displayed in Fig.~\ref{fig:test_conf} as box plot of p-values for different runs.
Since the distributions are similar despite the presence of noise, the p-values are high. 
This provides a reference for the p-values when accounting for noise.

To verify the validation procedure, two artificial verification inputs abbreviated with `RV' and `RV2' are constructed and compared with the original input. 
The first is simply deleting all objects in a scene. 
The second is to delete all vehicles which are within 2 m from the heading axis of the ego vehicle.
Both inputs are constructed to be implausible relevance criteria.
The latter filters out 5\% of the objects within the heuristic input region of the prediction network, thus filtering out fewer objects than the relevance criteria developed in this work.
The error distributions in Fig.~\ref{fig:distribution_errors} exhibit general visual similarity, all being larger than the prediction noise.
Nevertheless, the enlarged image indicates that the verification inputs differ from the case using all inputs. 
Testing for equality of distributions between verification and all inputs (A-RV and A-RV2) in Fig.~\ref{fig:test_conf} shows p-values which are orders of magnitude smaller than for A-A.
The large spread in p-values is a consequence of prediction noise.
Average p-values are below the exemplary threshold value of 0.005~\cite{Benjamin.2018}. 
This indicates a high confidence that the error distributions are different for the verification input and all inputs. 
Since the prediction is impacted by the verification input, the verification criteria are successfully identified as invalid. 

Similar to the verification inputs, the prediction is performed on the input filtered according to the relevance criteria of this work abbreviated as `R'.
The relevance criteria filter out 10\% of the objects included within the heuristic input region of the prediction network.
Fig.~\ref{fig:distribution_errors} shows that even with an enlarged image, the distributions appear visually similar. 
The results of testing for equality of all and of relevant inputs (A-R) are shown in Fig.~\ref{fig:test_conf}
The p-values are similar to A-A, indicating that any dissimilarity in error distributions is within the range caused by noise. 
Since the prediction performance is unaffected, the relevance criteria of this work are not falsified.
Accordingly, the validation results support the relevance criteria from this work. 

\section{Discussion}

This section discusses the results of both the validation method proposed as well as the validation of the extended urban relevance model.

\subsection{Validation Method}

The proposed validation method relies on a motion prediction network.
Compared to \gls{pkl}~\cite{Philion.2020b}, fewer requirements are imposed upon the motion prediction.
Discrete output trajectories and non-deterministic implementations are applicable without modification. 

The method explicitly considers the global performance of the prediction, showing the error to be in the order of 1 m. 
Notably, the global performance is robust to local noise which may appear in the prediction for a single scenario even with identical inputs. 
We suspect the inherent uncertainty of future trajectories causes stochastic behavior in the predictions. 
Since the underlying trajectory distribution is unknown, the prediction performance can only be assessed over a large number of scenarios. 
Conversely, this means that the local prediction performance in a single scenario is not meaningful since it cannot be disentangled from the stochastic component.
Therefore, the local stability of the prediction as used by \gls{pkl}~\cite{Philion.2020b} is not suited as relevance measure for some prediction networks.
This is visible in the noise results as exhibited by a non-deterministic prediction implementation as used in this work. 

The validation method of this work is successful in falsifying the verification input. 
The global performance of the prediction deteriorates, which agrees with the intuition that the vehicle directly in front is relevant to the vehicle. 
This verification succeeds despite the fact that only 5\% of the objects are filtered out, as opposed to the 10\% for the relevance criteria. 
However, further study is required to understand the impact of thresholds for p-values.   
In addition, further research is required to determine to what degree invalid samples can be resolved in large datasets. 
However, higher sensitivity of the validation may be possible if subsets of specific scenarios are considered individually. 

\subsection{Extended Urban Relevance Model}

Generally, we first develop analytic relevance criteria which are then reconciled with a prediction component at validation.
We consider this approach to have the following advantages.
Firstly, no complex neural networks are required when applying the relevance criteria. 
This is beneficial with regards to implementation effort~\cite{Wolf.2021} and computational resources.
Additionally, neural networks are prone to failures due to their lack of robustness~\cite{Houben.2022, Willers.2020} which may impact directly determining relevance.
However, the proposed approach is robust to these failures since the distribution across many samples is considered. 
Additionally, the analytic relevance criteria are fully interpretable.

These criteria are developed by applying an existing method to derive relevance.
The method is successfully applied to extend the results to the urban domain.
Applying the proposed validation method on a large-scale dataset shows that the prediction performance is unaffected by the relevance criteria proposed in this work. 
The validation is currently restricted to a single prediction network and one dataset.
For this case, the validity of the relevance criteria is supported. 
However, only 10\% of objects can be excluded from the data based on relevance. 
It is currently not known if the specificity of the relevance model can be improved while maintaining validity.

\section{Conclusion and Outlook}

This work presents \gls{sure}-Val, the extension of a recent method to derive relevance to the urban domain.
For this purpose, the methodology is applied to the intersection and the passing scenario.
Additionally, a novel method for validating analytic relevance criteria using a motion prediction is introduced. 
The relevance criteria and the validation method are applied on a public dataset using an exemplary motion prediction component. 
The verification of the validation procedure shows that the validation method itself is feasible.
At the same time, the validation results support the relevance criteria obtained in this work.
Additionally, the analytic relevance criteria are computationally efficient, interpretable and agnostic with regards to the implementation of downstream components.
We hope that both the relevance criteria as well as the validation method can serve as reference to consider these aspects more explicitly in the future.

For future research, it is desirable to further explore the limitations of the relevance criteria and their validation.
One aspect is the impact of the prediction architecture on relevance.
Another aspect is the consideration of datasets including more varied scenarios.
Further validation on such datasets which include unusual and dangerous driving situations is required to gain confidence in relevance criteria in general. 

\section{Acknowledgement}

The research leading to these results is funded by the German Federal Ministry for Economic Affairs and Climate Action within the project VVM - Verification Validation Methods under grant number 19A19002S, as well as by the German Federal Ministry of Education and Research within the project VIVID with the grant number 16ME0173 based on a decision of the Deutscher Bundestag. The authors would like to thank the consortia for the successful cooperation.


\bibliographystyle{plain}
\bibliography{Mo, St}



\end{document}